\documentclass[tablecaption=bottom,wcp]{jmlr} %

\jmlrproceedings{}{}

\title[AutoBayes]{AutoBayes: A Compositional Framework for Generalized Variational Inference}

\author{\Name{Toby St Clere Smithe} \addr VERSES Research
  \AND
  \Name{Marco Perin} \addr VERSES Research}

\usepackage{microtype}
\usepackage{enumitem}
\setlist{itemsep=0em, topsep=0em, parsep=0em}%

\usepackage{tikz}
\usepackage{tikzit}
\usetikzlibrary{cd, babel}

\newcommand*{\secref}[1]{\S\ref{#1}}

\def\kto{\rightsquigarrow}

\newcommand{\xkto}[1]{\overset{#1}{\kto}}
\def\xto{\xrightarrow}

\renewcommand{\d}{\mathrm{d}}

\def\pa{\mathrm{pa}}

\newcommand{\lenscirc}{
  \mathbin{\mathoverlap{\circ}{\raisebox{0.375ex}{\scalebox{1.0}[0.33]{$|$}}}}
}
\newcommand{\lensto}{\mathrel{\ooalign{\hfil$\mapstochar\mkern5mu$\hfil\cr$\to$\cr}}}

\makeatletter
\providecommand*{\xmapstofill@}{%
  \arrowfill@{\mapstochar\relbar}\relbar\rightarrow
}
\providecommand*{\xmapsto}[2][]{%
  \ext@arrow 0395\xmapstofill@{#1}{#2}%
}

\makeatletter
\newcommand{\mathoverlap}[2]{\mathpalette\mathoverlap@{{#1}{#2}}}
\newcommand{\mathoverlap@}[2]{\mathoverlap@@{#1}#2}
\newcommand{\mathoverlap@@}[3]{\ooalign{$\m@th#1#2$\crcr\hidewidth$\m@th#1#3$\hidewidth}}
\makeatother

\newcommand{\mcirc}{\mathbin{\mathoverlap{\circ}{\cdot}}} %
\newcommand*{\smallmcirc}{\raisebox{0.18ex}{\scalebox{0.66}{$\mcirc$}}}

\newcommand{\mto}{\mathoverlap{\rightarrow}{\smallmcirc\,}}

\usepackage{mathabx}
\usepackage{stmaryrd}
\newcommand\interp[1]{\ensuremath{\llbracket #1 \rrbracket}}

\newcommand{\gto}{\rightarrowtriangle}

\def\id{\mathsf{id}}

\tikzstyle{dot}=[inner sep=0.0mm, outer sep=0.0mm, minimum size=1mm, draw, shape=circle]
\tikzstyle{dot-5mm}=[dot, minimum size=5mm]
\tikzstyle{dot-1cm}=[dot, minimum size=1cm]
\tikzstyle{wcopy}=[dot, fill=white, scale=2.0]
\tikzstyle{bcopy}=[dot, fill=black, scale=2.0]
\tikzstyle{box}=[fill=white, draw=black, shape=rectangle]
\tikzstyle{box-5mm}=[box, minimum size=5mm, shape aspect=1]
\tikzstyle{box-7mm}=[box, minimum size=7mm, shape aspect=1]
\tikzstyle{box-1cm}=[box, minimum size=10mm, shape aspect=1]
\tikzstyle{dia}=[fill=white, draw=black, shape=diamond]
\tikzstyle{dia-5mm}=[dia, minimum size=5mm, shape aspect=1]
\tikzstyle{dia-7mm}=[dia, minimum size=7mm, shape aspect=1]
\tikzstyle{effect}=[regular polygon, regular polygon sides=3, draw]
\tikzstyle{state0}=[regular polygon, regular polygon sides=3, draw, shape border rotate=0]
\tikzstyle{state90}=[regular polygon, regular polygon sides=3, draw, shape border rotate=90]
\tikzstyle{state180}=[regular polygon, regular polygon sides=3, draw, shape border rotate=180]
\tikzstyle{state270}=[regular polygon, regular polygon sides=3, draw, shape border rotate=270]
\tikzstyle{scalar}=[diamond, draw, inner sep=1pt]
\tikzstyle{ground0}=[my ground, draw, inner sep=0pt, minimum width=4.2pt, minimum height=11.2pt, anchor=input, rotate=0]
\tikzstyle{ground90}=[my ground, draw, inner sep=0pt, minimum width=4.2pt, minimum height=11.2pt, anchor=input, rotate=90]

\input{strings.tikzdefs}

\ifexternalizetikz\tikzexternaldisable\fi
\newsavebox\sbground
\savebox\sbground{%
  \begin{tikzpicture}[baseline=0pt]
    \draw (0,-.1ex) to (0,.85ex)
    node[ground IEC,draw,anchor=input,inner sep=0pt,
    minimum width=3.15pt,minimum height=8.4pt,rotate=90] {};
  \end{tikzpicture}%
}

\newsavebox\sbcopy
\savebox\sbcopy{%
  \begin{tikzpicture}[baseline=0pt]
    \node[wcopy,scale=0.7] (a) at (0,3.8pt) {};
    \draw (a) -- +(-90:.21);
    \draw (a) -- +(45:.21);
    \draw (a) -- +(135:.21);
  \end{tikzpicture}}

\ifexternalizetikz\tikzexternalenable\fi

\newsavebox\bsbcopy
\savebox\bsbcopy{%
  \begin{tikzpicture}[baseline=0pt]
    \node[bcopy,scale=0.7] (a) at (0,3.8pt) {};
    \draw (a) -- +(-90:.21);
    \draw (a) -- +(45:.21);
    \draw (a) -- +(135:.21);
  \end{tikzpicture}}
\newcommand{\bcopy}{\mathord{\usebox\bsbcopy}}

\ifexternalizetikz\tikzexternalenable\fi

\def\Pa{\mathcal{P}}

\DeclareMathOperator*{\E}{\mathbb{E}}

\def\KL{\mathsf{KL}}
\def\VFE{\mathsf{VFE}}

\begin{document}

\maketitle

\begin{abstract}
  We introduce a new compositional framework for generalized variational inference, clarifying the different parts of a model, how they interact, and how they compose.
  We explain that both exact Bayesian inference and the loss functions typical of variational inference (such as variational free energy and its generalizations) satisfy chain rules akin to that of reverse-mode automatic differentiation, and we advocate for exploiting this to build and optimize models accordingly.
  To this end, we construct a series of compositional tools: for building models; for constructing their inversions; for attaching local loss functions; and for exposing parameters.
  Finally, we explain how the resulting \textit{parameterized statistical games} may be optimized locally, too.
  We illustrate our framework with a number of classic examples, pointing to new areas of extensibility that are revealed.
\end{abstract}

\section{Introduction}

Key to the success and vibrancy of deep learning is automatic differentiation (\textit{autodiff}) and particularly its \textit{reverse mode}, which allows the backpropagation of loss through arbitrary differentiable programs.
In turn, key to automatic differentiation is the chain rule from calculus (and, in reverse mode, its transpose), by which the derivatives of composite functions may be computed from the derivatives of their parts.
Mathematically, the chain rule follows from the functoriality of the (co)tangent bundle structure; in reverse mode, for differentiable maps $X\xto{f}Y\xto{g}Z$, we have $\d_x(g\circ f)(z) = \d_xf\circ\d_{f(x)}g(z)$, for any cotangent vector $z$ to $Z$ and point $x$ in $X$.

This story is of course well known.
Less well known is that there is an analogous chain rule for Bayesian inference \citep{Braithwaite2023Compositional}, related to (but different from) the better known chain rule for entropy.
The existence of this rule means that, to invert a complex model, it suffices to compose the inversions of its factors.
And just as reverse mode automatic differentiation yielded a renaissance in differential machine learning, we believe that this chain rule for inference heralds a new era for Bayesian methods.
Moreover, this new era can be built upon the advances in optimization due to deep learning, by focusing our efforts on variational methods.
To this end, we show that the loss functions adopted in variational inference exhibit a compositional structure that is compatible with the Bayesian chain rule.
Our work is therefore not only of theoretical interest, specifying as it does the computational structure of a powerful new framework for generalized variational inference.

In order to carve nature cleanly and precisely at its joints, we make use of the mathematical language of category theory.
This allows us: to separate the specification of a statistical model from that of its inversion; to enable models to make use of \textit{dependent types}; to define and compose loss functions locally while ensuring correctness; to distinguish the different roles that likelihood and regularization play; to annotate models with the families of distribution that they may yield; to separate syntax (model specification) from semantics (optimization); and to ensure that all of these components interact well with each other.
All the same, we have done our best to minimize the demands of novel mathematics in the main text of this paper.

Their authors being unaware of all this structure, existing frameworks are only able to approximate it.
For example, various probabilistic programming languages (PPLs) allow the user to separate the specification of the model from its inversion (typically called a `guide' \citep{Ritchie2016Deep,Bingham2019Pyro}), but there is typically only a weak structural coupling between these two \citep{Pham2024Programmable}.
Moreover, because PPLs are typically aimed at describing and sampling from arbitrary distributions, the machinery involved in computing inversions is typically the same sampling procedure used for computing the models themselves: the weak coupling between the two means that the compositional structure cannot be fully exploited.

In the variational setting, two conceptual frameworks get closer to the mark: the ``generalized variational inference'' of Knoblauch \textit{et al} \citep{Knoblauch2019Generalized}, and the ``Bayesian learning rule'' of Khan \textit{et al} \citep{Khan2023Bayesian}.
These works are notable for considering explicitly the structure of the loss functions involved in variational inference, but due to their focus on optimizing single (parametric families of) models rather than models in general, they miss the gains to be had from considering models themselves as compositional.
This single-model focus of much of the statistical machine learning literature produces an excess of manual work, as researchers derive loss functions for complex models by hand.
We advocate instead using modern mathematical and computational tools to automate this work, and thus improve modularity and reusability.

\paragraph{Contributions}
\begin{enumerate}
\item We describe a compositional framework for specifying probabilistic models, generalizing Bayesian networks and enabling models with dependent types.
  Our framework does not supplant PPLs: models can be composed of probabilitistic programs.
\item We make explicit the relationship between the specification of a model and its inversion, the compositional structure of which is given by the Bayesian chain rule.
\item We explain precisely how loss functions for complex models are obtained from simple parts, separating the likelihood terms from the differently-behaved regularizers.
\item We clearly separate the `syntax' of model specification from the `semantics' of optimization, as there are often multiple algorithms applicable to each model type.
\item To that end, we clarify how to annotate models with the specific families of distributions that they yield, observing that this expressivity is traded for compositionality.
\item Finally, optimization algorithms need parameters to act on, so we explain how these parameters may be exposed coherently with respect to the rest of the structure.
\end{enumerate}
Although we leave the implementation of the framework for future work, we demonstrate its utility by exhibiting a number of examples in Appendix \secref{apdx:examples}.

\paragraph{Notation}

We write $f:X\to Y$ to denote deterministic maps (functions), and $c:X\kto Y$ to denote stochastic maps (measure kernels).
The latter yield measures for each element of their domain, which we write in conditional probability style, as $c(\d y|x)$.
Thus, we denote sets and spaces with upper-case letters and elements of those spaces with corresponding lower-case letters.
We will assume that each measurable space $X$ is equipped with a canonical measure, denoted $\d x$.
When a kernel $c:X\kto Y$ is associated with a density function with respect to this canonical measure, we will denote it by $p_c(y|x)$, so that $c(\d y|x) = p_c(y|x)\,\d y$.

We use $\circ$ and $\bullet$ to denote the composition of functions and kernels respectively, with the latter defined by the Chapman-Kolmogorov equation: $X\xkto{c}Y\xkto{d}Z$ is given by $(d\bullet c)(\d z|x) = \int_{y:Y} d(\d z|y) \, c(\d y|x)$.
We use $1$ to denote a singleton set, and note that measures are equivalent to kernels with domain $1$.
The pushforward of a measure $\pi$ along a kernel $c$ is written $c_*\pi$ and defined by composition: $c_*\pi = c\bullet\pi$.
The Bayesian inversion of a kernel $c$ with respect to a prior $\pi$ is a kernel (almost surely unique) in the opposite direction, which we denote $c^\dag_\pi:Y\kto X$.
We write $[x = y]$ to denote the indicator function defined as $1$ when $x=y$ and $0$ otherwise.

\section{Beyond Bayesian Networks}

A \textit{Bayesian network} is a distribution over a product of spaces $X_v$ indexed by the nodes $v\in V$ of a directed acyclic graph $G$ that factors according to the graph structure: informally, $p(x) = \prod_{v\in V} p(x_v|x_{\pa(v)})$, where $x\in \prod_{v\in V} X_v$, $x_v \in X_v$, and $x_{\pa(v)} \in \prod_{v\in\pa(v)} X_v$ with $\pa(v)$ denoting the set of parents of a node $v\in V$.
Thus, Bayesian networks describe the factorization structure of joint distributions.
In applications, Bayesian networks are typically presented as `fully-formed' objects, to aid readers' understanding of the models that they encode: often, this \textit{graphical model} is depicted alongside a symbolic expression of the joint distribution.
When the model is also instantiated in code, this may be similarly monolithic.

This state of affairs misses much useful structure of such networks: they are mathematically well behaved \textit{compositional} objects, mechanically reusable and recombinable, suitable to the kind of analysis and transformation that compilers apply to traditional programming languages.
It is on this basis that we build our AutoBayes framework.

\begin{definition}[Open model]
  If $X$ and $Y$ are measurable spaces, then an \textnormal{open model} $p:X\mto Y$ consists of a pair of a measurable space $\interp p$ and a measure kernel $p:X\kto \interp p\times Y$.
  We call the domain $X$ of the open model the \textnormal{unobserved space}, the codomain $Y$ the \textnormal{observed space}, and $\interp p$ the \textnormal{latent space}.
  If $\interp p \cong 1$, we call $p$ a \textnormal{pure model}; if $X \cong \interp p \cong 1$, we call $p$ a \textnormal{pure distribution}; and if $X \cong 1$ but $\interp p \ncong 1$, we call $p$ a \textnormal{joint distribution}.
\end{definition}

\begin{remark}
  We say `open' to emphasize that these models are open to composition and are therefore not quite `complete': they correspond to conditional distributions.
  Open models with domain $1$ correspond to non-conditional distributions, and one may say that an open model $1\mto 1$ is a `closed' model.
  We are careful to distinguish observed, unobserved, and latent spaces, and in doing so we emphasize that our models are inherently directed.
  We think of the observed space as ``where the observable data lives'', with the unobserved space standing in for what in other literature is sometimes called the parameter or latent variable.
  We will soon see how both the unobserved and observed spaces may become latent in our sense; and later we will have a precise notion of \textit{parameter}, too.
  Finally, we say ``model'' rather than ``Bayesian network'', as our models will be more general than Bayesian networks.
\end{remark}

\begin{remark}
  We wrote in the introduction that our framework does not supplant PPLs, even though it is intended to be computationally implemented.
  The reason for this is that AutoBayes exists at a higher level: because probabilistic programs define (s-finite) kernels \citep{Staton2017Commutative}, they may be used to define open models (and their inversions), which can in turn be composed and optimized using the AutoBayes framework.
\end{remark}

\begin{definition}[Composition of models]
  Given open models $p:X\mto Y$ and $q:Y\mto Z$, we can compose them to form an open model $q\mcirc p:X\mto Z$.
  The composite latent space $\interp{q\mcirc p}$ is $\interp p\times Y \times \interp q$ and the composite kernel $q\mcirc p:X\kto \interp p\times Y\times \interp q\times Z$ is defined by $(q\mcirc p)(\d s, \d y, \d t, \d z|x) = q(\d t, \d z|y)\, p(\d s,\d y|x)$.
\end{definition}

Here, we see that the latent space is used to hold the parts of models that become `hidden' when we compose them.
This captures the compositional behavior of joint distributions: note that if $p$ and $q$ are pure models, then $(q\mcirc p)(\d y, \d z|x) = q(\d z|y)\,p(\d y|x)$ is not pure.
In particular, if $p$ is a pure distribution and $q$ a pure model, then $q\mcirc p$ is a joint distribution.

\begin{remark}
  That this composition is well-behaved is verified by the fact that it yields a bicategory whose 1-cells are open models.
  This follows from \citet[\S5.2.1]{Smithe2023Mathematical} and \citet[\S2]{Smithe2024Copycomposition}.
  The identity model $\id_X:X\mto X$ is pure, defined by the Dirac kernel $\id_X(\d x|x') = [x = x'] \, \d x$.
\end{remark}

As a Bayesian network, we can depict $q\mcirc p:1\mto Y\mto Z$ in the familiar way on the left below.
But category theory proposes an alternative, more expressive, depiction, on the right.
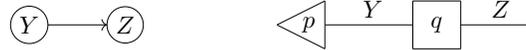
\[
\scalebox{0.85}{\begin{tikzpicture}
	\begin{pgfonlayer}{nodelayer}
		\node [style=dot-5mm] (0) at (-1.5, 0) {$\;Y\;$};
		\node [style=dot-5mm] (2) at (1.5, 0) {$\;Z\;$};
	\end{pgfonlayer}
	\begin{pgfonlayer}{edgelayer}
		\draw [->] (0) to (2);
	\end{pgfonlayer}
\end{tikzpicture}
}
\qquad\qquad
\scalebox{0.85}{\begin{tikzpicture}
	\begin{pgfonlayer}{nodelayer}
		\node [style=none] (3) at (-3, 0) {$p$};
		\node [style=none] (4) at (-2.5, 0.75) {};
		\node [style=none] (5) at (-4, 0) {};
		\node [style=none] (6) at (-2.5, -0.75) {};
		\node [style=none] (7) at (-2.5, 0) {};
		\node [style=none] (8) at (1, 0) {$q$};
		\node [style=none] (9) at (0.25, 0.75) {};
		\node [style=none] (10) at (1.75, 0.75) {};
		\node [style=none] (11) at (0.25, -0.75) {};
		\node [style=none] (12) at (1.75, -0.75) {};
		\node [style=none] (13) at (0.25, 0) {};
		\node [style=none] (14) at (1.75, 0) {};
		\node [style=none] (23) at (-1, 0.5) {$Y$};
		\node [style=none] (24) at (3, 0.5) {$Z$};
		\node [style=none] (25) at (4, 0) {};
	\end{pgfonlayer}
	\begin{pgfonlayer}{edgelayer}
		\draw (5.center) to (4.center);
		\draw (4.center) to (6.center);
		\draw (6.center) to (5.center);
		\draw (7.center) to (13.center);
		\draw (9.center) to (10.center);
		\draw (10.center) to (12.center);
		\draw (12.center) to (11.center);
		\draw (11.center) to (9.center);
		\draw (14.center) to (25.center);
	\end{pgfonlayer}
\end{tikzpicture}
}
\]
The \textit{string diagram} on the right has the advantage of depicting both the spaces ($Y$, $Z$) and the distributions ($p$, $q$) that together form the model, and it makes explicit which parts of the model are pure distributions (\textit{e.g.}, `priors') and which are conditional: the former are depicted with triangular boxes and only outgoing edges; the latter are squares with both outgoing and incoming edges.
The dangling edges represent the openness of the model, and we can compose diagrams along compatible edges, and read off precisely the form of the composite model: here, we have composed the model $q$ after $p$.
Thus this depiction also allows us to distinguish open from closed (\textit{e.g.} those models with priors from those without).

As well as sequentially, we can also compose models in parallel, by what is sometimes called the `tensor' product.

\begin{definition}[Parallel composition]
  If $q:Y\mto Z$ and $q':Y'\mto Z'$ are two open models, then there is an open model $q\otimes q':Y\otimes Y'\mto Z\otimes Z'$ defined as follows.
  Let $Y\otimes Y'$ denote the product space $Y\times Y'$ (and similarly for $Z\otimes Z'$).
  The latent space $\interp{q\otimes q'}$ is $\interp q\times\interp{q'}$.
  The kernel $q\otimes q':Y\times Y'\kto\interp{q}\times\interp{q'}\times Z\times Z'$ is defined by $(q\otimes q')(\d t,\d t',\d z,\d z'|y,y') = q(\d t,\d z|y)\,q'(\d t',\d z'|y')$.
\end{definition}

\begin{remark}
  This tensor defines a monoidal product on the bicategory of open models, whose unit is the space $1$.
\end{remark}

Using the tensor, we can represent models involving products of distributions.
A classic Bayesian network of this kind is depicted on the left below; a corresponding string diagram in our formalism is on the right.
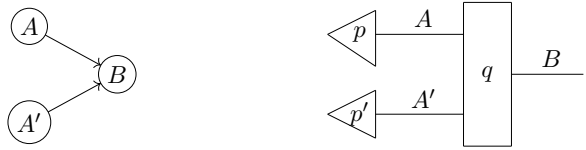
\[
\scalebox{0.85}{\begin{tikzpicture}
	\begin{pgfonlayer}{nodelayer}
		\node [style=dot-5mm] (0) at (-1.25, -1.5) {$\;A'\;$};
		\node [style=dot-5mm] (2) at (1.5, 0) {$\;B\;$};
		\node [style=dot-5mm] (3) at (-1.25, 1.5) {$\;A\;$};
	\end{pgfonlayer}
	\begin{pgfonlayer}{edgelayer}
		\draw [->] (0) to (2);
		\draw [->] (3) to (2);
	\end{pgfonlayer}
\end{tikzpicture}
}
\qquad\qquad\qquad
\scalebox{0.85}{\begin{tikzpicture}
	\begin{pgfonlayer}{nodelayer}
		\node [style=none] (3) at (-3, 1.25) {$p$};
		\node [style=none] (4) at (-2.5, 2) {};
		\node [style=none] (5) at (-4, 1.25) {};
		\node [style=none] (6) at (-2.5, 0.25) {};
		\node [style=none] (7) at (-2.5, 1.25) {};
		\node [style=none] (8) at (1, 0) {$q$};
		\node [style=none] (9) at (0.25, 2.25) {};
		\node [style=none] (10) at (1.75, 2.25) {};
		\node [style=none] (11) at (0.25, -2.25) {};
		\node [style=none] (12) at (1.75, -2.25) {};
		\node [style=none] (13) at (0.25, 1.25) {};
		\node [style=none] (14) at (1.75, 0) {};
		\node [style=none] (23) at (-1, 1.75) {$A$};
		\node [style=none] (24) at (3, 0.5) {$B$};
		\node [style=none] (25) at (4, 0) {};
		\node [style=none] (26) at (-3, -1.25) {$p'$};
		\node [style=none] (27) at (-2.5, -0.5) {};
		\node [style=none] (28) at (-4, -1.25) {};
		\node [style=none] (29) at (-2.5, -2) {};
		\node [style=none] (30) at (-2.5, -1.25) {};
		\node [style=none] (31) at (0.25, -1.25) {};
		\node [style=none] (32) at (-1, -0.75) {$A'$};
	\end{pgfonlayer}
	\begin{pgfonlayer}{edgelayer}
		\draw (5.center) to (4.center);
		\draw (4.center) to (6.center);
		\draw (6.center) to (5.center);
		\draw (7.center) to (13.center);
		\draw (9.center) to (10.center);
		\draw (10.center) to (12.center);
		\draw (12.center) to (11.center);
		\draw (11.center) to (9.center);
		\draw (14.center) to (25.center);
		\draw (28.center) to (27.center);
		\draw (27.center) to (29.center);
		\draw (29.center) to (28.center);
		\draw (30.center) to (31.center);
	\end{pgfonlayer}
\end{tikzpicture}
}
\]
The string diagram depicts the model $q\mcirc(p\otimes p'):1\mto B$.
Note that $A\otimes A'$ is latent.

Because the latent space is just a factor of the codomain of the kernel making up an open model, it can be revealed as part of the observed space by a purely formal manoeuvre; this corresponds to a 2-cell in the bicategory of open models.
Thus, we obtain $\mathsf{reveal}_{A\times A'}\bigl(q\mcirc(p\otimes p')\bigr)$ as a pure distribution $1\mto A\otimes A'\otimes B$.

We can also extend any model $q:X\mto Y$ to include a ``dummy variable'' corresponding to any space $A$, by tensoring with the corresponding identity model, as in $\id_A\otimes q:A\otimes X\mto A\otimes Y$, which we abbreviate to $A\otimes q$.
This is useful when we want to allow information to ``flow past a factor'', as in the model depicted below right (corresponding Bayesian network left):
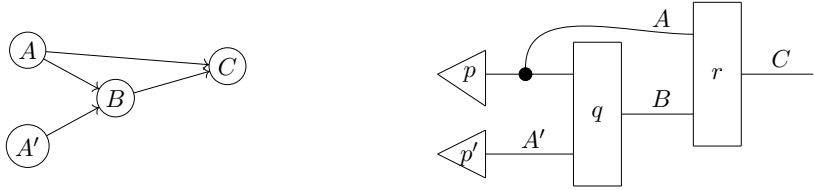
\[
\scalebox{0.85}{\begin{tikzpicture}
	\begin{pgfonlayer}{nodelayer}
		\node [style=dot-5mm] (0) at (-3.25, -1.5) {$\;A'\;$};
		\node [style=dot-5mm] (2) at (-0.5, 0) {$\;B\;$};
		\node [style=dot-5mm] (3) at (-3.25, 1.5) {$\;A\;$};
		\node [style=dot-5mm] (4) at (3, 1) {$\;C\;$};
	\end{pgfonlayer}
	\begin{pgfonlayer}{edgelayer}
		\draw [->] (0) to (2);
		\draw [->] (3) to (2);
		\draw [->] (3) to (4);
		\draw [->] (2) to (4);
	\end{pgfonlayer}
\end{tikzpicture}
}
\qquad\qquad\qquad
\scalebox{0.85}{\begin{tikzpicture}
	\begin{pgfonlayer}{nodelayer}
		\node [style=none] (3) at (-4.75, 0.75) {$p$};
		\node [style=none] (4) at (-4.25, 1.5) {};
		\node [style=none] (5) at (-5.75, 0.75) {};
		\node [style=none] (6) at (-4.25, -0.25) {};
		\node [style=none] (7) at (-4.25, 0.75) {};
		\node [style=none] (8) at (-0.75, -0.5) {$q$};
		\node [style=none] (9) at (-1.5, 1.75) {};
		\node [style=none] (10) at (0, 1.75) {};
		\node [style=none] (11) at (-1.5, -2.75) {};
		\node [style=none] (12) at (0, -2.75) {};
		\node [style=none] (13) at (-1.5, 0.75) {};
		\node [style=none] (14) at (0, -0.5) {};
		\node [style=none] (23) at (1.25, 2.5) {$A$};
		\node [style=none] (24) at (1.25, 0) {$B$};
		\node [style=none] (25) at (2.25, -0.5) {};
		\node [style=none] (26) at (-4.75, -1.75) {$p'$};
		\node [style=none] (27) at (-4.25, -1) {};
		\node [style=none] (28) at (-5.75, -1.75) {};
		\node [style=none] (29) at (-4.25, -2.5) {};
		\node [style=none] (30) at (-4.25, -1.75) {};
		\node [style=none] (31) at (-1.5, -1.75) {};
		\node [style=none] (32) at (-2.75, -1.25) {$A'$};
		\node [style=none] (33) at (3, 0.75) {$r$};
		\node [style=none] (34) at (2.25, 3) {};
		\node [style=none] (35) at (3.75, 3) {};
		\node [style=none] (36) at (2.25, -1.5) {};
		\node [style=none] (37) at (3.75, -1.5) {};
		\node [style=none] (38) at (2.25, 2) {};
		\node [style=none] (39) at (3.75, 0.75) {};
		\node [style=none] (40) at (5, 1.25) {$C$};
		\node [style=none] (41) at (6, 0.75) {};
		\node [style=none] (42) at (2.25, -0.5) {};
		\node [style=bcopy] (43) at (-3, 0.75) {};
	\end{pgfonlayer}
	\begin{pgfonlayer}{edgelayer}
		\draw (5.center) to (4.center);
		\draw (4.center) to (6.center);
		\draw (6.center) to (5.center);
		\draw (9.center) to (10.center);
		\draw (10.center) to (12.center);
		\draw (12.center) to (11.center);
		\draw (11.center) to (9.center);
		\draw (14.center) to (25.center);
		\draw (28.center) to (27.center);
		\draw (27.center) to (29.center);
		\draw (29.center) to (28.center);
		\draw (30.center) to (31.center);
		\draw (34.center) to (35.center);
		\draw (35.center) to (37.center);
		\draw (37.center) to (36.center);
		\draw (36.center) to (34.center);
		\draw (39.center) to (41.center);
		\draw (7.center) to (43);
		\draw (43) to (13.center);
		\draw [in=180, out=90] (43) to (38.center);
	\end{pgfonlayer}
\end{tikzpicture}
}
\]
The $\bcopy$ symbol in the string diagram denotes a `copier', which ensures that the $A$ inputs to $q$ and $r$ are equal to the $A$ output from $p$.
As a kernel $A\kto A\times A$, this is given by $\bcopy(\d a_1, \d a_2|a_0) = [a_1 = a_0 = a_2] \, \d a_1 \, \d a_2$.

With all these ingredients, it is possible to show that \textit{all} Bayesian networks can be represented as the composition of open models.
This follows from \citet[Theorem 4.5]{Fong2013Causal} and \citet[\S2]{Smithe2024Copycomposition}.
Specifically, if $\{X_i\}_{i\in V}$ is a Bayesian network, it can be written as an open model $p:1\mto\otimes_{i\in V} X_i$ obtained by composing factors of the form $p_i:\otimes_{j\in\pa(i)} X_j \mto X_i$ for each $i\in V$.\footnote{
First, sort the nodes such that $j<i$ iff there is no directed path from $i$ to $j$.
Associate to each node $X_i$ an open model $p_i:\otimes_{j\in\pa(i)} X_j \mto X_i$ representing the associated conditional distribution.
Then, reveal the parents (so they are accessible to later factors), and extend with dummy variables corresponding to all those $j<i$ not in $\pa(i)$ --- \textit{i.e.}, form $\left(\otimes_{j<i, j\notin\pa(i)} X_j\right)\otimes \mathsf{reveal}_{\pa(i)}(p_i):\otimes_{j<i} X_j\mto\otimes_{j\leq i}X_j$.
Next, compose these factors in sequence following the ordering.}

The open models framework encompasses more than Bayesian networks, however.
First, assuming we allow unnormalized measures, we can relax the acyclicity restraint.
This means we can represent models as on the right below, which corresponds to a joint distribution of the form $[a_1 = a_2]\,r(\d a_2,\d c|b)\,q(\d b|a_1) \, \d a_1$, and which could only be represented as an ambiguous generalized Bayesian network of the form on the left below.
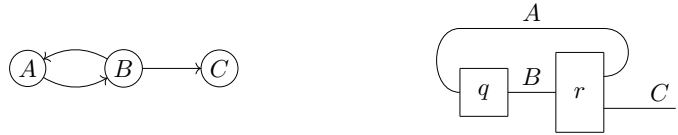
\[
\scalebox{0.85}{\begin{tikzpicture}
	\begin{pgfonlayer}{nodelayer}
		\node [style=dot-5mm] (2) at (0, 0) {$\;B\;$};
		\node [style=dot-5mm] (3) at (-3, 0) {$\;A\;$};
		\node [style=dot-5mm] (4) at (3, 0) {$\;C\;$};
	\end{pgfonlayer}
	\begin{pgfonlayer}{edgelayer}
		\draw [->, bend right] (3) to (2);
		\draw [->] (2) to (4);
		\draw [->, bend right] (2) to (3);
	\end{pgfonlayer}
\end{tikzpicture}
}
\qquad\qquad\qquad
\scalebox{0.85}{\begin{tikzpicture}
	\begin{pgfonlayer}{nodelayer}
		\node [style=none] (5) at (-2, -0.75) {$q$};
		\node [style=none] (6) at (-2.75, 0) {};
		\node [style=none] (7) at (-1.25, 0) {};
		\node [style=none] (8) at (-2.75, -1.5) {};
		\node [style=none] (9) at (-1.25, -1.5) {};
		\node [style=none] (11) at (-1.25, -0.75) {};
		\node [style=none] (12) at (-0.5, 1.75) {$A$};
		\node [style=none] (13) at (-0.5, -0.25) {$B$};
		\node [style=none] (14) at (0.25, -0.75) {};
		\node [style=none] (22) at (1, -0.75) {$r$};
		\node [style=none] (23) at (0.25, 0.5) {};
		\node [style=none] (24) at (1.75, 0.5) {};
		\node [style=none] (25) at (0.25, -2) {};
		\node [style=none] (26) at (1.75, -2) {};
		\node [style=none] (31) at (0.25, -0.75) {};
		\node [style=none] (32) at (1.75, -1.25) {};
		\node [style=none] (34) at (-2.75, -0.75) {};
		\node [style=none] (35) at (1.75, -0.25) {};
		\node [style=none] (36) at (4, -1.25) {};
		\node [style=none] (37) at (3.5, -0.75) {$C$};
		\node [style=none] (38) at (1.75, 1.25) {};
		\node [style=none] (39) at (-2.75, 1.25) {};
	\end{pgfonlayer}
	\begin{pgfonlayer}{edgelayer}
		\draw (6.center) to (7.center);
		\draw (7.center) to (9.center);
		\draw (9.center) to (8.center);
		\draw (8.center) to (6.center);
		\draw (11.center) to (14.center);
		\draw (23.center) to (24.center);
		\draw (24.center) to (26.center);
		\draw (26.center) to (25.center);
		\draw (25.center) to (23.center);
		\draw (32.center) to (36.center);
		\draw (34.center)
			 to [in=-180, out=-180, looseness=1.25] (39.center)
			 to (38.center)
			 to [in=0, out=0, looseness=1.75] (35.center);
	\end{pgfonlayer}
\end{tikzpicture}
}
\]
Formally, this composite involves two important pure models, a `cup' $1\mto A\otimes A$ and a `cap' $A\otimes A\mto 1$, both of which enforce an equality constraint.
The cup is defined as the distribution $\mathsf{cup}_A(\d a_1, \d a_2) = [a_1 = a_2] \, \d a_1 \, \d a_2$, and the cap is defined as the function $\mathsf{cap}_A(a_1, a_2) = [a_1 = a_2]$.\footnote{
An unnormalized kernel $X\kto 1$ is equivalent to a function $X\to[0,\infty)$.}
Note that the cup turns an unobserved space into an observed one, and the cap \textit{vice versa}.
This will be useful in Example \ref{ex:sup-learn} (supervised learning).

\begin{remark}
  The existence of well-behaved cups and caps makes the bicategory of open models \textnormal{self-dual compact closed}.
\end{remark}

The second generalization that our framework enables is to models of \textit{dependent} type, but, for reasons of space, we relegate this discussion to Appendix \secref{apdx:dep-typ}.
Thus, having established our basic framework, let us now use it for inference.

\section{Local Inversions for Compositional Models}

Given a measure kernel $c:X\kto Y$, its Bayesian inversion is a function $c^\dag:\Pa X\to\{Y\kto X\}$, where $\Pa X$ denotes the space of (s-finite) measures on $X$ and $\{Y\kto X\}$ is the set of kernels $Y\kto X$.
Applying $c^\dag$ to a prior $\pi\in\Pa X$ yields a kernel $c^\dag_\pi:Y\kto X$ in the opposite direction to $c$, defined canonically by
\[ c^\dag_\pi(\d x|y) = \frac{c(\d y|x)\,\pi(\d x)}{(c_*\pi)(\d y)} \]
for all $y$ in the support of $c_*\pi$.
This expression is Bayes' law, and $c^\dag_\pi$ is called the exact posterior with respect to $\pi$.

The chain rule for Bayesian inference \citep{Braithwaite2023Compositional} states that, given $c:X\kto Y$ and another kernel $d:Y\kto Z$, the inversion of the composite $d\bullet c$ is the (inverse) composite of the inversions.
That is, $(d\bullet c)^\dag_\pi = c^\dag_\pi\bullet d^\dag_{c_*\pi}$.
Note the formal similarity to the reverse-mode chain rule from calculus, $\d_x(g\circ f) = \d_xf\circ \d_{f(x)}g$.
Despite its simple statement, its easy verification (just use Bayes' law!), and its utility, the Bayesian chain rule is surprisingly ill-known.

One reason for this may be that the composition of measure kernels by pushforward (\textit{i.e.}, integration) is expensive, and so not often given much consideration in the context of approximate inference.
However, it is also easily verified that, if $c$ and $d$ are instead \textit{open models} (as in the preceding section), and we define $c^\dag:\Pa X\to\{Y\mto X\}$ by mapping $\pi\in\Pa X$ to the open model whose kernel is the inversion of that of $c$ at $\pi$, then $(d\mcirc c)^\dag_\pi = c^\dag_\pi\mcirc d^\dag_{c_*\pi}$ also.
That is, \textit{open models satisfy the Bayesian chain rule}, too.
And the composition of open models does not involve integration.
(Moreover, we will see below that open models constitute the right framework for composing free energies.)

The key advance that follows from recognizing the Bayesian chain rule is that it allows us to define inversions \textit{locally}, and still construct correct `global' posteriors.
Moreover, it tells us the structure that a posterior must have in order to be a correct posterior for a complex model.
Of course, these local inversions do not have to be exact: we can still compose approximate posteriors following the chain rule, and the composite inversion that results will be structured correctly by construction.
This line of thinking motivates the definition of \textit{Bayesian lenses}: models paired with corresponding local inversions.

\begin{definition}[Bayesian lens]
  A \textnormal{Bayesian lens} $X\lensto Y$ consists of a pair $(c,c')$ of an open model $c:X\mto Y$ and a function $c':\Pa X\to\{Y\kto X\times\interp{c}\}$ mapping priors on $X$ to kernels inverse to $c$.
\end{definition}

\begin{definition}[Exact Bayesian lens]
  Given an open model $c:X\mto Y$ there is a canonical \textnormal{exact} Bayesian lens $(c,c^\dag)$ where $c^\dag$ is defined by the generalization of Bayes' law to open models: given a prior $\pi\in\Pa X$ (and ignoring questions of support),
  \[ c^\dag_\pi(\d x,\d a|y) = \frac{c(\d a,\d y|x)\,\pi(\d x)}{\int_{a':\interp{c}} (c_*\pi)(\d a',\d y)} \; . \]
\end{definition}

\begin{remark}
  In other literature, less focused on the composition of models, the forwards part $c(\d y|x)$ is often simply written as $p(y|x)$, a prior simply as $p(x)$, an exact inversion with respect to that prior as $p(x|y)$, and an inexact inversion as $q(x|y)$, or even simply as $q(x)$.
  We believe our notation is less overloaded and so clarifies the different parts of a model.
\end{remark}

\begin{definition}[Composition of Bayesian lenses]
  Given $(c,c'):X\lensto Y$ and $(d,d'):Y\lensto Z$, their composite $(d,d')\diamond(c,c'):X\lensto Z$ is defined as $(d\mcirc c, c'\lenscirc d'_c)$, where $c'\lenscirc d'_c$ is the function $\Pa X\to\{Z\kto X\times\interp{d\mcirc c}\}$ mapping $\pi\in\Pa X$ to the kernel defined by
  \[ \left(c'_\pi\lenscirc d'_{c_*\pi}\right)(\d x,\d a,\d y,\d b|z) = c'_\pi(\d x, \d a|y) \, d'_{c_*\pi}(\d y,\d b|x) \; . \]
\end{definition}

\begin{theorem}[Chain rule for open models]
  Define a function $(-)^\dag$ mapping open models $c:X\mto Y$ to exact Bayesian lenses $(c)^\dag := (c,c^\dag):X\lensto Y$.
  This function satisfies $(d\mcirc c)^\dag = (d,d^\dag)\diamond(c,c^\dag) = (d)^\dag\diamond(c)^\dag$.
  That is to say, $(-)^\dag$ is \textnormal{functorial}.
\end{theorem}

\begin{remark}
  The notion of Bayesian lens here is a generalization of that of \citet{Braithwaite2023Compositional} to the open models case, and the chain rule theorem is a generalization likewise.
  The definition of composition $\diamond$ yields a bicategory of Bayesian lenses (with the obvious choice of identity lenses), and so the chain rule theorem establishes that $(-)^\dag$ is a pseudofunctor between the bicategory of open models and the bicategory of these Bayesian lenses.\footnote{
  Strictly speaking, one needs a technical adjustment to the notion of Bayesian lens to account for the possibility that inversions may not be fully supported, and even then inversions are only defined up to almost sure equality, so that $(-)^\dag$ is only almost surely a pseudofunctor.}
\end{remark}

The parallel composition extends from open models to Bayesian lenses; but we must be careful with the inversions.

\begin{definition}[Parallel composition]
  The tensor $(c,c')\otimes(d,d')$ of $(c,c'):X\lensto Y$ and $(d,d'):X'\lensto Y'$ is defined as $(c\otimes d, c'\boxtimes d')$, where $c'\boxtimes d'$ is the function $\Pa(X\times X')\to\left\{Y\times Y'\kto X\times X'\times\interp{c\otimes d}\right\}$ mapping $\omega\in\Pa(X\times X')$ to the kernel $(c'\boxtimes d')_\omega$ given by
  \[ (c'\boxtimes d')_\omega(\d x,\d x',\d a,\d a'|y,y') = c'_{\omega_X}(\d x,\d a|y) \, d'_{\omega_{X'}}(\d x',\d a'|y') \]
  where $\omega_X$ is the $X$-marginal and $\omega_{X'}$ the $X'$-marginal of $\omega$.
\end{definition}

\begin{remark} \label{rmk:lax-tensor}
  This definition entails that composing inversions in parallel is lossy, because the inversions being composed can only accept the marginals of a joint prior.
  Consequently, $(-)^\dag$ is only a so-called \textit{lax} monoidal functor: $(c\otimes d)^\dag \neq c^\dag\otimes d^\dag$ (with a 2-cell witnessing the inequality).
\end{remark}

Like open models, Bayesian lenses also have cups and caps, allowing us to bend unobserved spaces into observed ones (and vice versa), and represent cyclic models (and their inversions).
$\mathsf{cup}_A:1\lensto A\otimes A$ is defined as $(\mathsf{cup}_A,\mathsf{cap}_A)$, \textit{i.e.} with the cup forward and the (constant function on the) cap as the inversion.
The cap of Bayesian lenses is defined dually, with the cap model forward and the cup as its inversion.

\section{Composing Complex Loss Functions}

Bayesian lenses enable the compositional specification of models equipped with local inversions (posteriors)\footnote{
Note that there is nothing forcing any level of granularity on the inversions: one can build a complex open model compositionally, and then only attach an inversion to the composite; or one can attach inversions to the factors of the complex model, and compose all the parts together as lenses.}.
In general, the local inversions can be quite different from the exact inversions defined by Bayes' law, and, in practice, they will often be specified by parametric families (often denoted $\mathcal{Q}$ in the literature).
The important question for a practitioner of approximate inference is then: how good are these approximate inversions?
How close do they get to the exact posterior?

In the context of variational inference, this question is usually answered at first by a divergence on the space of distributions in question, typically the Kullback-Leibler divergence (or relative entropy).
For example, given a Bayesian lens $(c,c'):X\lensto Y$, one can define a function $\Pa X\times Y\to[0,\infty]$ by $(\pi,y)\mapsto D_{KL}\left(c'_\pi(y),c^\dag_\pi(y)\right)$, and then minimize this function for a given prior $\pi$ and dataset in $Y$.
Denoting this function by $\KL(c,c')$, one notes that it is parametric in the lens; \textit{i.e.}, it could just as easily have been defined as $\KL(d,d')$ for $(d,d'):Y\lensto Z$.

This leads one to wonder whether there is a chain rule for the relative entropy, corresponding to the chain rule for exact inference, and, of course, there is: it is quite easy to show that $\KL\bigl((d,d')\lenscirc(c,c')\bigr)(\pi,z) = \E_{(y,b)\sim d'_{c_*\pi}(z)}\bigl[\KL(c,c')(\pi,y)\bigr] + \KL(d,d')(c_*\pi,z)$; \citep[\S5.3.3.1]{Smithe2023Mathematical}.
But the relative entropy is just a starting point for variational inference, and one hopes that there might be a general framework for attaching local losses to local inversions that captures the gamut of loss functions used in practice.
It is such a framework that we establish now.

Although one might want to minimize the relative entropy, this quantity still depends on evaluating the computationally intractable $c^\dag_\pi$.
Therefore, in variational inference, one typically optimizes a bound on the relative entropy.
The classic choice of bound is the quantity called the \textit{variational free energy}\footnote{
Variational free energy is known elsewhere as the ``evidence upper bound'' (EUBO) or negative ``evidence lower bound'' (negative ELBO).},
which is the sum of the relative entropy and the (negative) marginal log likelihood.
This can also be defined as a function parametric in lenses, but the addition of the likelihood term breaks the strong compositionality of the relative entropy \citep[\S5.3.3.3]{Smithe2023Mathematical}.
It is for this reason that we must be a bit cleverer with our set-up.

\begin{definition}[Variational free energy]
  Let $(c,c'):X\lensto Y$ be a Bayesian lens.
  Its variational free energy $\VFE(c,c')$ is the function $\Pa X\times Y\to[0,\infty]$ defined by $\VFE(c,c')(\pi,y) = \KL(c,c')(\pi,y) - \log p_{c_Y\bullet\pi}(y)$, where $c_Y$ is the $Y$-marginal of the kernel $c$, so that $(c_Y\bullet\pi)(\d y) = \int_{a:\interp c} (c_*\pi)(\d a,\d y)$.
  (When $c$ is a pure model, $p_{c_Y\bullet\pi}(y) = p_{c_*\pi}(y)$.)
\end{definition}

The reason that this is a useful bound on the relative entropy is that the log likelihood interacts with the relative entropy to eliminate the direct dependence on the exact inversion.

\begin{proposition}[Alternative forms of $\VFE$]
  By expanding the definition of $\KL$, we can write $\VFE$ in the following ways; note the lack of $c^\dag_\pi$ in the second form:
  \begin{align}
    \VFE(c,c')(\pi,y)
    &= \E_{(x,a)\sim c'_\pi(y)} \left[ \log p_{c'_\pi}(x,a|y) - \log p_{c^\dag_\pi}(x,a|y) \right] - \log p_{c_Y\bullet\pi}(y) \\
    &= \E_{(x,a)\sim c'_\pi(y)} \left[ \log p_{c'_\pi}(x,a|y) - \log p_c(a,y|x) - \log p_\pi(x) \right] \\
    &= \E_{(x,a)\sim c'_\pi(y)} \left[ - \log p_c(a,y|x) - \log p_\pi(x) \right] - H\left(c'_\pi(y)\right) \label{eq:VFE}
  \end{align}
  where $H\left(c'_\pi(y)\right) = D_{KL}\left(c'_\pi(y), \d x\otimes\d a\right)$ is the Shannon entropy of $c'_\pi(y)$.
\end{proposition}

The third form above explains why $\VFE$ is called as it is: the \textit{Helmholtz} free energy is a difference between expected energy and entropy; so the term inside the expectation is sometimes also called the \textit{energy} of the model.
This form of $\VFE$ is the key to the loss-function part of the AutoBayes framework:
the central observation is that the energy and entropy parts behave differently.
Importantly, the energy term in $\VFE(c,c')(\pi,y)$ has a contribution from both $c$ and the prior $\pi$.
We will see that energies compose by simple addition, but entropies (and thus losses built from them) compose like the chain rule.

\begin{remark}
  Observations like these also seem to underlie the proposals of \citet{Knoblauch2019Generalized} for `generalized' variational inference, and the proposals of \citet{Khan2023Bayesian} for a ``Bayesian learning rule''.
  Both sets of authors observe that typical learning objectives (such as ELBOs or free energies) can be written as the sum of a loss or likelihood term, in expectation under some posterior, plus a divergence or entropy term regularizing this posterior.
  But neither set notices the important compositional implications that follow.
\end{remark}

Building on these observations, our next step is to associate to Bayesian lenses the corresponding loss function terms: a `likelihood' or \textit{energy} term on the one hand; and a `regularizer' or \textit{entropy} term on the other.
The optimization target will then be obtained by combining these.

\begin{definition}[Statistical game]
  A \textnormal{statistical game} $c:X\gto Y$ consists of a quadruple $(c,c',l^c,H^c)$, where $(c,c')$ is a Bayesian lens $X\lensto Y$, $l^c$ is a function $X\times \interp{c}\times Y\to[0,\infty]$ and $H^c$ is a function $\Pa X\times Y\to[0,\infty]$.
  We call $l^c$ the \textnormal{energy} or \textnormal{likelihood}, and $H^c$ the \textnormal{entropy} or \textnormal{regularizer}.
  We combine $l^c$ and $H^c$ into a \textnormal{loss} (a generalized free energy) $F^c:\Pa X\times Y\to[0,\infty]$ by defining
  \[ F^c(\pi,y) = \E_{(x,a)\sim c'_\pi(y)} \left[ l^c(x,a,y) \right] - H^c(\pi,y) \; . \]
\end{definition}

\begin{remark}
  The concept of ``statistical game'' was originally introduced in \citet{Smithe2023Mathematical} and \citet{Smithe2023Approximate}, but those definitions don't yield the well-behaved compositional structure that we present here; the novelty of our definition is the correct decomposition of the free energy into energy and entropy terms, and the acknowledgment of their different compositional behavior.
  The term ``statistical game'' makes reference to game theory, as loss functions can be seen as utility or fitness functions, and compositional game theory is also built on a framework of (different) lenses \citep{Ghani2018Compositional}.
\end{remark}

\begin{definition}[Composition of statistical games]
  Given statistical games $c:X\gto Y$ and $d:Y\gto Z$, their composite $d\diamond c:X\gto Z$ is defined as follows.
  Its Bayesian lens is given by the composition of the associated Bayesian lenses $(d,d')\diamond(c,c'):X\lensto Z$.
  The energy $l^{dc}:X\times\interp{c}\times Y\times\interp{d}\times Z\to[0,\infty]$ is defined by $l^{dc}(x,a,y,b,z) := l^c(x,a,y) + l^d(y,b,z)$.
  The entropy $H^{dc}:\Pa X\times Z\to[0,\infty]$ is defined by
  \[ H^{dc}(\pi,z) := \E_{(y,b)\sim d'_{c_*\pi}(z)}\left[H^c(\pi,y)\right] + H^d(c_*\pi,z) \; . \]
\end{definition}
A consequence of this composition law is that we obtain a chain rule for generalized free energies, analogous to that for the relative entropy cited above; when the energies are negative marginal log likelihoods and the entropies are Shannon, this specializes to a chain rule for the variational free energy.

\begin{theorem}[Chain rule for free energy] \label{thm:chain-vfe}
The loss $F^{dc}:\Pa X\times Z\to[0,\infty]$ satisfies
\begin{align*}
  F^{dc}(\pi,z)
  &= \E_{(x,a,y,b)\sim(c'_\pi\lenscirc d'_{c_*\pi})(z)} \left[ l^{dc}(x,a,y,b,z) \right] - H^{dc}(\pi,z) \\
  &= \E_{(y,b)\sim d'_{c_*\pi}(z)} \left[ \E_{(x,a)\sim c'_\pi(y)} \left[ l^c(x,a,y) \right] - H^c(\pi,y) + l^d(y,b,z) \right] - H^d(c_*\pi,z) \\
  &= \E_{(y,b)\sim d'_{c_*\pi}(z)} \left[ F^c(\pi,y) \right] + F^d(c_*\pi,z)
\end{align*}
\end{theorem}
This result is at the heart of the AutoBayes framework: it means that optimizing a complex model can be reduced to optimizing its parts, as long as the appropriate information (priors and posteriors) is passed forward and back correctly; it is a ``Bayesian autodiff'' framework.
Perhaps most importantly, this means that it is unnecessary to derive complex loss functions monolithically by hand, as is often done in statistical machine learning: they may be \textit{composed} mechanistically and locally instead, just like gradients in differentiable programming.
\begin{remark}
  An important sanity check on the structure is that the composition of statistical games (as with open models and Bayesian lenses) yields a bicategory.
  The identity game $X\gto X$ is given by the identity lens $X\lensto X$ equipped with constantly $0$ energy and entropy.
\end{remark}

The reader may have noted that, for a given pure statistical game $c:X\gto Y$ whose loss is the log marginal likelihood $-\log p_c(y|x)$ and whose entropy is the Shannon entropy $H(c'_\pi(y))$, the induced loss $F^c(\pi,y)$ is not precisely the variational free energy of equation \eqref{eq:VFE} above, for $F^c(\pi,y) = \E_{x\sim c'_\pi(y)} [-\log p_c(y|x)] - H(c'_\pi(y))$; it is missing the term $-\log p_\pi(x)$.
The reason for this is that $c$ is an \textit{open} model, and so this is an ``open free energy''.

Given a prior distribution $\pi:1\mto X$ (represented as an open model), we can turn it into a statistical game by first equipping it with a trivial inversion, then setting $l^\pi(x) = -\log p_\pi(x)$, and finally noting that $H^\pi(\cdot,x)$ must be $0$ (as the inversion is trivial).
Composing $c:X\gto Y$ after the resulting $\pi:1\gto X$ yields a composite statistical game with free energy $F^{c\pi}(\cdot,y) = \VFE(c,c')(\pi,y)$, as desired.
This shows that free energies really are compositional objects.

We end this section by sketching the parallel composition structure.

\begin{definition}[Parallel composition]
  The tensor $c\otimes d$ of $c:X\gto Y$ and $d:X'\gto Y'$ is defined as the tensor of the corresponding lenses equipped with the energy $l^{c\otimes d}:X\times X'\times\interp{c}\times\interp{d}\times Y\times Y'\to[0,\infty]$ and entropy $H^{c\otimes d}:\Pa(X\times X')\times Y\times Y'\to[0,\infty]$ given respectively by $l^{c\otimes d}(x,x',a,a',y,y') = l^c(x,a,y) + l^d(x',a',y')$ and $H^{c\otimes d}(\omega,y,y') = H^c(\omega_X,y) + H^d(\omega_{X'},y')$.
\end{definition}

\begin{remark}
  In the case that the entropies are Shannon, the laxness of the tensor (see Remark \ref{rmk:lax-tensor}) is measured by the mutual information.
\end{remark}

\section{Optimization via Functorial Semantics}

Just as an automatic differentiation system is somehow incomplete without an optimizer, and although our focus in this paper has been on composing Bayesian inversions, it would be remiss not to discuss the optimization of the associated loss functions.
Effectively, we think of the framework developed so far as compositional `syntax' for variational inference; we will also need `semantics'.

The first step on this road is to choose what precisely is to be optimized, as a loss function of the form $F^c(\pi,y)$ exposes no obvious parameter.
We want our framework to be agnostic about the role of the parameter (for instance, it could represent the weights of a neural network in the forward pass, or the natural parameter of an exponential family posterior), and so we make the following simple definition, which allows any part of a statistical game to depend on a parameter.

\begin{definition}
  A \textnormal{parameterized} statistical game $X\gto Y$ is a pair $(\Theta,c)$ of a space $\Theta$ and a function $c:\Theta\to\{X\gto Y\}$ from $\Theta$ to the set of statistical games $X\gto Y$.
  We may write each of the components of the game as $c(\d y|x;\theta)$, $c'_\pi(\d x|y;\theta)$, $l^c(x,y;\theta)$ and $H^c(\pi,y;\theta)$.
\end{definition}

With this definition, we have something that can be optimized.
Assuming that the function $c$ is differentiable, we can form the free energy and compute its gradients, $\nabla_\theta F^c(\pi,y;\theta)$.
In many applications, $\Theta$ will parameterize a statistical manifold, and thus be equipped with the Fisher information metric.
Gradient descent of generalized free energy with respect to this metric is what Khan and Rue \citep{Khan2023Bayesian} call the \textit{Bayesian learning rule}\footnote{
For Khan and Rue, $\Theta$ typically picks the natural parameter of the posterior.
But this means that $c'$ is for them not a map $\Theta\times\Pa X\times Y\to\Pa X$ but merely $\Theta\to\Pa X$.
The compositional structure of models therefore does not feature in their treatment.},
and it is this kind of gradient descent that we take to be the standard semantics for AutoBayes.

The questions for us are whether and how we can construct this semantics compositionally, using only the local loss function data, which amount to the `functoriality' of gradient descent.
First, we need to understand how the parameters interact with the composition.

\begin{definition}[Composition of parameterized statistical games]
  Given parameterized statistical games $(\Theta,c):X\gto Y$ and $(\Phi,d):Y\gto Z$, their composite is defined to be $(\Phi\times\Theta,d\blackdiamond c)$ where $d\blackdiamond c$ is now a function mapping $(\phi,\theta)$ to the statistical game $d(\phi)\diamond c(\theta)$.
  The parallel composition is defined similarly.
\end{definition}

And, crucially, we need to understand how the gradients are composed.

\begin{definition}[Composition of gradients]
  Given parameterized statistical games $(\Theta,c):X\gto Y$ and $(\Phi,d):Y\gto Z$, their gradients $\nabla_\theta F^c$ and $\nabla_\phi F^d$ may be composed to form $\begin{pmatrix} \nabla_\phi F^d(c(\theta)_*\pi,z;\phi) & \E_{y\sim d'_{c(\theta)_*\pi}(z;\phi)} \left[ \nabla_\theta F^c(\pi,y;\theta) \right] \end{pmatrix}^T$.
\end{definition}

The question of functoriality then reduces to whether this expression is equal to $\nabla_{\phi,\theta} F^{dc}$ when evaluated at the same points.
Looking back at Theorem \ref{thm:chain-vfe}, we see that, in general, it is not: the $F^d$ term may depend on $\theta$ (via the pushforward prior), and the $F^c$ term may depend on $\phi$ (via $y$).
However, this just means that the assignment of gradients is `lax', which can be accounted for mechanistically by an implementation.

\begin{remark}[Formalities]
  As before, parameterized statistical games form a monoidal bicategory.
  One then couples parameter spaces to their tangent bundles, and allows for maps back into those bundles.
  This step yields a fibration over parameterized statistical games, and the assignment of gradients to parameterized games is a lax section of this fibration.
\end{remark}

Let us end this paper with a brief discussion about how this gradient descent may be implemented, for it is in the algorithmics that much of the art of inference appears; and it is with the implementation of this framework that we concern ourselves with next.

First, we note that in a composite game, priors are propagated by pushforward (\textit{i.e.}, marginalization), and computing these is similarly expensive to computing exact inversions, so an approximation scheme is warranted here, too.
For this purpose, a simple algorithmic choice is to use belief propagation or variational message passing, and this can be shown to fit into the framework we have presented here.

Similarly, information propagates backwards by sampling from, or taking expectations under, the sequence of posteriors.
This is mathematically the correct thing to do, but again it implies computational difficulties: particularly, for example, if one is optimizing the parameters of the posteriors themselves.
It is for this reason that Khan \textit{et al} advocate the use of conjugate models where possible \citep{Khan2017ConjugateComputation,Khan2023Bayesian}: in this situation, many of these computations are greatly simplified. %
However, conjugate models require priors to be in particular families of distributions, and pushing forwards does not generally preserve these families, so approximations are again needed (\textit{e.g.} moment-matching, to project the priors back to the desired family).
To encode these families of distributions into the statistical game data (so that an implementation may make use of it) requires annotating the games with predicates---which can be done compositionally, but laxly.

The principal difficulties are thus in computing the (expectations involved in the) gradients, and different models call for different strategies.
And just as we had exact and approximate inversions above, we have exact and approximate gradients here.
For this reason, we expect that these different strategies---including the Laplace method, and delta rule, and sampling schemes of various kinds---will correspond to different ``semantics functors'': different compositional assignments of gradients to parameterized statistical games.

\bibliography{references}

\begin{thebibliography}{13}
\providecommand{\natexlab}[1]{#1}
\providecommand{\url}[1]{\texttt{#1}}
\expandafter\ifx\csname urlstyle\endcsname\relax
  \providecommand{\doi}[1]{doi: #1}\else
  \providecommand{\doi}{doi: \begingroup \urlstyle{rm}\Url}\fi

\bibitem[Bingham et~al.(2019)Bingham, Chen, Jankowiak, Obermeyer, Pradhan,
  Karaletsos, Singh, Szerlip, Horsfall, and Goodman]{Bingham2019Pyro}
Eli Bingham, Jonathan~P. Chen, Martin Jankowiak, Fritz Obermeyer, Neeraj
  Pradhan, Theofanis Karaletsos, Rohit Singh, Paul Szerlip, Paul Horsfall, and
  Noah~D. Goodman.
\newblock Pyro: Deep universal probabilistic programming.
\newblock \emph{J. Mach. Learn. Res.}, 20\penalty0 (1):\penalty0 973--978,
  January 2019.
\newblock ISSN 1532-4435.

\bibitem[Braithwaite et~al.(2023)Braithwaite, Hedges, and
  St~Clere~Smithe]{Braithwaite2023Compositional}
Dylan Braithwaite, Jules Hedges, and Toby St~Clere~Smithe.
\newblock The {{Compositional Structure}} of {{Bayesian Inference}}, May 2023.

\bibitem[Fong(2013-01-26, 2013)]{Fong2013Causal}
Brendan Fong.
\newblock \emph{Causal {{Theories}}: {{A Categorical Perspective}} on
  {{Bayesian Networks}}}.
\newblock PhD thesis, University of Oxford, 2013-01-26, 2013.

\bibitem[Ghani et~al.(2018)Ghani, Hedges, Winschel, and
  Zahn]{Ghani2018Compositional}
Neil Ghani, Jules Hedges, Viktor Winschel, and Philipp Zahn.
\newblock Compositional {{Game Theory}}.
\newblock In \emph{Proceedings of the 33rd {{Annual ACM}}/{{IEEE Symposium}} on
  {{Logic}} in {{Computer Science}}}, pages 472--481, Oxford United Kingdom,
  July 2018. ACM.
\newblock ISBN 978-1-4503-5583-4.
\newblock \doi{10.1145/3209108.3209165}.

\bibitem[Khan and Lin(2017)]{Khan2017ConjugateComputation}
Mohammad~Emtiyaz Khan and Wu~Lin.
\newblock Conjugate-{{Computation Variational Inference}} : {{Converting
  Variational Inference}} in {{Non-Conjugate Models}} to {{Inferences}} in
  {{Conjugate Models}}, April 2017.

\bibitem[Khan and Rue(2023)]{Khan2023Bayesian}
Mohammad~Emtiyaz Khan and H{\aa}vard Rue.
\newblock The {{Bayesian Learning Rule}}.
\newblock \emph{Journal of Machine Learning Research}, 24\penalty0
  (281):\penalty0 1--46, September 2023.

\bibitem[Knoblauch et~al.(2019)Knoblauch, Jewson, and
  Damoulas]{Knoblauch2019Generalized}
Jeremias Knoblauch, Jack Jewson, and Theodoros Damoulas.
\newblock Generalized {{Variational Inference}}.
\newblock December 2019.

\bibitem[Pham et~al.(2024)Pham, Wang, Saad, and Hoffmann]{Pham2024Programmable}
Long Pham, Di~Wang, Feras~A. Saad, and Jan Hoffmann.
\newblock Programmable {{MCMC}} with {{Soundly Composed Guide Programs}}.
\newblock \emph{Proceedings of the ACM on Programming Languages}, 8\penalty0
  (OOPSLA2):\penalty0 1051--1080, October 2024.
\newblock ISSN 2475-1421.
\newblock \doi{10.1145/3689748}.

\bibitem[Ritchie et~al.(2016)Ritchie, Horsfall, and Goodman]{Ritchie2016Deep}
Daniel Ritchie, Paul Horsfall, and Noah~D. Goodman.
\newblock Deep {{Amortized Inference}} for {{Probabilistic Programs}}, October
  2016.

\bibitem[St~Clere~Smithe(2023{\natexlab{a}})]{Smithe2023Approximate}
Toby St~Clere~Smithe.
\newblock Approximate {{Inference}} via {{Fibrations}} of {{Statistical
  Games}}.
\newblock In \emph{Electronic {{Proceedings}} in {{Theoretical Computer
  Science}}}, volume 397, pages 279--298, College Park, MD, June
  2023{\natexlab{a}}. Open Publishing Association.
\newblock \doi{10.4204/EPTCS.397.17}.

\bibitem[St~Clere~Smithe(2023{\natexlab{b}})]{Smithe2023Mathematical}
Toby St~Clere~Smithe.
\newblock \emph{Mathematical {{Foundations}} for a {{Compositional Account}} of
  the {{Bayesian Brain}}}.
\newblock {{DPhil}}, University of Oxford, 2023{\natexlab{b}}.

\bibitem[St~Clere~Smithe(2024)]{Smithe2024Copycomposition}
Toby St~Clere~Smithe.
\newblock Copy-composition for probabilistic graphical models, June 2024.

\bibitem[Staton(2017)]{Staton2017Commutative}
Sam Staton.
\newblock Commutative {{Semantics}} for {{Probabilistic Programming}}.
\newblock In \emph{Programming {{Languages}} and {{Systems}}}, pages 855--879.
  Springer Berlin Heidelberg, 2017.
\newblock \doi{10.1007/978-3-662-54434-1_32}.

\end{thebibliography}

\appendix

\section{Examples} \label{apdx:examples}

\begin{example}[Gaussian mixture model / maximum likelihood estimation]
  Here, we have an unobserved finite set $M$ and a Gaussian distribution $c$ over observed $Y$ conditional on $M$, along with a prior distribution $\pi$ over $M$.
  The marginal distribution $c_*\pi$ on $Y$ is thus a mixture of Gaussians, and the prior may be parameterized by the probabilities assigned to each ``mixture component'' (each element of $M$).
  The aim of the game is then to optimize these probabilities in order to maximize the marginal likelihood on $Y$, given data $y$.
  We turn $c$ and $\pi$ into statistical games $M\gto X$ and $1\gto M$ as follows.
  For simplicity, we equip $c$ with its exact inversion $c^\dag$, yielding a lens $(c,c^\dag)$, and we let $l^c$ and $H^c$ be given by the negative log-likelihood and entropy respectively.
  To construct the game $1\gto M$, we note that the inversion (and thus entropy) are trivial, and let $l^\pi$ again be given by the negative log-likelihood.
  Under these circumstances, $F^{c\pi}(\cdot,y) = -\log p_{c_*\pi}(y)$.
  We expose no parameter on $c$, but we let $\pi:1\gto M$ be parameterized by the mixing probabilities, so that the function $\pi:\Pa M\to\{1\gto M\}$ simply maps $\alpha$ to $(\alpha,\cdot, -\log p_\alpha,0)$.
  Thus, descending the $\alpha$-gradient of $F^{c\pi}(\cdot,y;\alpha)$ is the same as maximizing the likelihood with respect to the mixing probabilities.
\end{example}

\begin{example}[Expectation-maximization]
  Expectation-maximization is typically used to compute a maximum likelihood estimate of the parameter of a model involving some unobserved component.
  Thus suppose we have a lens $c:X\lensto Y$ and a prior lens $\pi:1\lensto X$ (with trivial inversion).
  We equip them both with negative log-likelihoods for energies, and $0$ entropies, yielding games $c:X\gto Y$ and $\pi:1\gto X$.
  Then $F^{c\pi}(\cdot, y) = \E_{x\sim c'_\pi(y)}[-\log p(x,y)]$, where $p(x,y) = p_c(y|x)\, p_\pi(x)$ is the density of the joint model.
  Computing this constitutes the \textnormal{expectation} step of the EM algorithm.
  Now suppose the composite model is parameterized in $\Theta$.
  Maximizing $F^{c\pi}(\cdot, y; \theta)$ with respect to $\theta\in\Theta$ corresponds to the \textnormal{maximization} step of the algorithm.
\end{example}

\begin{example}[Variational Bayesian expectation-maximization]
  In VBEM, one extends the preceding example so that $\Theta$ forms part of the model, with games $c:\Theta\times X\gto Y$ and $\pi:\Theta\gto X$, which can be composed to form a game $c\pi:\Theta\gto Y$ by $\theta\mapsto c(\theta,-)\diamond\pi(\theta)$.
  One also then has a prior $\psi$ on $\Theta$, which we can take to be parameterized in $\Psi$.
  This induces a parameterized game $(\Psi,\psi):1\gto\Theta$ which can be composed with $(1,c\pi):\Theta\gto Y$.
  VBEM amounts to performing gradient descent on the resulting composite loss with respect to the parameter $\Psi$.
\end{example}

\begin{example}[Supervised learning] \label{ex:sup-learn}
  We can use the `cups' introduced above to incorporate supervised learning into the framework:  in this context, one typically knows the `unobserved' labels in $X$ corresponding to a set of observed data in $Y$, and a parameterized model $c:X\gto Y$ to be trained accordingly.
  By composing the tensor $X\otimes c:X\otimes X\gto X\otimes Y$ after the cup $1\gto X\otimes X$, one obtains a model in which both $X$ and $Y$ are `observed'.
  In this case, the inversion $c'$ will typically trivialize (as the prior is given by a deterministic sample), although the regularizer $H^c$ may not.
  The resulting loss $F$ will then only depend on the parameter and the paired data, and may be optimized accordingly.
\end{example}

\begin{example}[Bayesian deep learning]
  The previous example may of course be extended, much as EM was extended to VBEM above: we may consider a model $c:\Theta\otimes X\gto Y$ and only apply the cup to $X$, yielding a model $\Theta\gto X\otimes Y$ to be optimized.
  For instance, consider that the forwards component of $c$ may be a neural network with weights in $\Theta$, and making a stochastic prediction of the data in $Y$; the energy $l^c(\theta,x,y)$ may then be a complex loss function associated with a corresponding machine learning model.
  In Bayesian deep learning, one typically has not only a stochastic model of the data, but also a prior on the weights $\Theta$, which may easily be incorporated in this framework as a game $1\gto \Theta$.
  If the inversion $c'$ is `mean-field' factorized into a product over $\Theta$ and $X$ independently, then the cup trivializes the factor over $X$, leaving only a posterior over $\Theta$.
  The classic situation in Bayesian deep learning then corresponds to optimizing the parameters of this posterior --- but note that, in this case, both $c$ and the prior may themselves be complex models constructed compositionally.
\end{example}

\section{On dependently typed models} \label{apdx:dep-typ}

Usually, a joint distribution $p(\d x, \d y)$ is a distribution over a product space $X\times Y$, which can be understood as ``$X$-many copies of $Y$''.
Thus, for every $x$, $y$ is an element of $Y$.
This is a `simply' typed model.
In a \textit{dependently typed} model, the type of $y$ (\textit{i.e.}, the particular space $Y$) may depend on the choice of $x\in X$.
For example, at $x_1$, we might have $y\in\mathbb{R}^2$, but at a different $x_2$, we might have $y\in\mathbb{R}^3$.
Thus, instead of the product space $X\times Y$, the joint distribution is over the dependent sum (disjoint union) space $\sum_{x\in X} Y_x$.

This can be useful in modelling.
Consider a global weather report model: given coordinates for a location, it returns the expected weather at that location.
But the `type' of a weather report may differ depending on location: at sea, it might include information about tides and wind that is irrelevant or undefined on land.
If such a model were learned, enforcing this structure using types would save the model having to infer it from data.

A similar story can be told for conditional distributions.
Usually, a conditional distribution $q(\d y|x)$ is formalized as a kernel $X\kto Y$.
But in the dependent case, the codomain type depends on the specific value of the domain, which we might write as $(x:X)\kto Y_x$.
Formally, this corresponds to a \textit{stochastic section} of the projection $\sum_{x\in X} Y_x \to X$; \textit{i.e.}, a kernel $X\kto \sum_{x\in X} Y_x$ that preserves $X$.
Stochastic sections thus allow us to extend the open models formalism to dependent types.

\end{document}